# Automated Pavement Crack Segmentation Using U-Net-Based Convolutional Neural Network

**STEPHEN L. H. LAU**[ID][1,2], **EDWIN K. P. CHONG**[ID][3], **(Fellow, IEEE),**
**XU YANG**[1,4], **(Member, IEEE), AND XIN WANG**[ID][2], **(Senior Member, IEEE)**
[1]School of Highway, Chang'an University, Xi'an 710064, China
[2]School of Engineering, Monash University Malaysia, Subang Jaya 47500, Malaysia
[3]Department of Electrical and Computer Engineering, Colorado State University, Fort Collins, CO 80523, USA
[4]Department of Civil Engineering, Monash University, Clayton, VIC 3800, Australia

Corresponding authors: Xin Wang (wang.xin@monash.edu) and Xu Yang (xu.yang@monash.edu)

This work was supported in part by the 111 Project of Sustainable Transportation for Urban Agglomeration in Western China under Grant B200035, and in part by the Australia Research Council (ARC) Hub under Grant IH180100010.

**ABSTRACT** Automated pavement crack image segmentation is challenging because of inherent irregular patterns, lighting conditions, and noise in images. Conventional approaches require a substantial amount of feature engineering to differentiate crack regions from non-affected regions. In this paper, we propose a deep learning technique based on a convolutional neural network to perform segmentation tasks on pavement crack images. Our approach requires minimal feature engineering compared to other machine learning techniques. We propose a U-Net-based network architecture in which we replace the encoder with a pretrained ResNet-34 neural network. We use a ''one-cycle'' training schedule based on cyclical learning rates to speed up the convergence. Our method achieves an $F1$ score of 96% on the CFD dataset and 73% on the Crack500 dataset, outperforming other algorithms tested on these datasets. We perform ablation studies on various techniques that helped us get marginal performance boosts, i.e., the addition of spatial and channel squeeze and excitation (SCSE) modules, training with gradually increasing image sizes, and training various neural network layers with different learning rates.

**INDEX TERMS** Convolutional neural network, deep learning, fully convolutional network, pavement crack segmentation, U-Net.

## I. INTRODUCTION

Crack formation on pavements poses a safety hazard to road users. The principal causes of pavement crack formation include traffic, moisture, and construction quality [1]. The quality of roads worsens with time owing to wear and tear. Continual traffic flow in urban areas exacerbates this problem. A study done in 2006 revealed that accidents due to road conditions in the United States alone cost $217.5 billion [2]. The risk increases with road usage, and the consequence can be as severe as death. The maintenance of pavements is thus a priority to ensure the safety of road users.

In many pavement crack systems, road images are collected by performing line-scan or area-scan cameras. A few examples include the DHDV detection system [3] used in the American expressway, the PAVUE system [4] used in Sweden, and the GERPHO system [5] used in France. One way to improve pavement maintenance is by using computational algorithms to segment the pavement cracks from the background. If it achieves a reasonable accuracy, then it can usefully be deployed by pavement engineers to replace the traditional method of visually inspecting the pavement cracks before repairing them.

Image processing techniques such as thresholding [6], mathematical morphology [7], and edge detection [8] are often used to segment cracks in non-destructive testing (NDT), but are very sensitive to noise in images and cannot generalize the differentiation of a crack from its background.

In this paper, we propose a deep learning algorithm for pavement crack segmentation, which we demonstrate to outperform other methods. Our deep learning algorithm reduces the need for manual feature extraction because it can learn the essential features required to classify each pixel as a crack or not. Traditional approaches typically require the selection of essential features in each given image. This process can be cumbersome and subjective [9]. Deep learning algorithms can often forego this manual step and learn these features automatically. Therefore, once the algorithm is fully trained, the detection of any crack would take a much shorter time.

We summarize our contributions as follows:

- We introduce transfer learning by incorporating a pretrained neural network in our proposed neural network architecture to achieve better performance on benchmark datasets.
- We introduce the concurrent spatial and channel squeeze and excitation (SCSE) modules in our proposed network architecture, which increase the performance, as we show in the ablation studies.
- We introduce the use of various techniques, including training with progressively increasing image sizes and training various layers in the neural network with different learning rates, which also helped to obtain marginally better performances, respectively.

The rest of this paper is organized as follows: Section II discusses the previous work on pavement crack image segmentation. In Section III, we explain the architecture of our convolutional neural network, the loss function we minimize, and the various steps of the training procedure. In Section IV, we provide prediction results on benchmark datasets and the quantitative metrics we use to evaluate the results. Besides, we also compare the performance of our algorithm with other methods. In Section V, we provide ablation studies among various algorithm designs to evaluate their influence on the performance. Finally, Section VI concludes the paper by summarizing our work while pointing out its limitations and some ways to improve it.

## II. RELATED WORK

This section briefly reviews recent work on the segmentation of pavement crack images. Oliveira and Correia [6] segmented crack from images by preprocessing images using morphological filters, then applying dynamic thresholding to filter out the darker pixels. In [10], Ai *et al.* constructed a probability map for the crack prediction based on pixel intensity and multi-scale neighborhood information. Shi *et al.* [11] proposed a crack detection framework based on random forests, heavily relying on feature extraction of an annotated image database. Li *et al.* [12] extracted candidate cracks using a windowed intensity path-based method before segmenting the cracks using a crack evaluation model based on a multivariate statistical hypothesis test.

For deep learning methods, Bang *et al.* [13] used a deep residual network as the encoder of the neural network, and a fully connected network as its decoder. Fan *et al.* [14] proposed a convolutional neural network (CNN) method for structural prediction as a multi-label problem. They extracted small image patches from images and labeled each patch as *positive* if the center pixel is a crack pixel, and *negative* otherwise. The prediction output corresponds to a $5 \times 5$ image. Despite yielding excellent results, the method suffers from long inference times per testing image. Zhang *et al.* [15] used CNN for segmentation. Their approach involves a feature extractor to provide input to a second CNN. The same authors also used a recurrent neural network on 3D pavement images to predict the local paths with the highest probability to form crack patterns [16]. These methods are not end-to-end networks and still require feature extraction. Yang *et al.* [17] proposed a novel architecture that involves feature pyramid and hierarchical boosting modules. The feature pyramid modules merge feature maps from two successive CNN layers in the downsampling blocks, whereas the hierarchical boosting modules assign weights to easy and hard samples accordingly. For pavement crack segmentation tasks, several researchers, such as Jenkins *et al.* [18] and Nguyen *et al.* [19], used a U-Net-based network architecture in their supervised learning algorithms. However, these methods did not use transfer learning in their network architecture design, which works well in computer vision tasks [20]. In our paper, we propose using a pretrained neural network as part of our U-Net-based network architecture, while also introducing various training techniques to increase the performance further.

## III. PROPOSED METHOD

The method we propose involves a supervised learning algorithm in which we optimally fit a function approximator $f : \boldsymbol{x} \rightarrow \boldsymbol{y}$ using training data to map images $\boldsymbol{x}$ to their respective labels $\boldsymbol{y}$. In deep learning, this function approximator takes the form of a parameterized neural network comprising multiple layers of weights and biases. The neural network is trained to minimize a loss function $L = g(\boldsymbol{x}, \boldsymbol{y}, \boldsymbol{\theta})$ by tuning the parameters $\boldsymbol{\theta}$. This tuning is done with a gradient descent-based optimizer.

### A. NETWORK ARCHITECTURE

The network architecture we propose is a U-Net-based architecture [21] with a ResNet-34 [22] encoder, which was pretrained on ImageNet (with its last two layers, i.e., pooling layer and classifier removed), as shown in Fig. 1. (Our description assumes some background in convolutional neural networks.) This fully convolutional network receives a three-channel (RGB) image as input and produces a one-channel (binary) output of the same size. The ResNet-34 encoder begins with a convolutional layer with a kernel size of $7 \times 7$ and stride of 2. This is followed by a batch normalization (BN) layer [23], a rectified linear unit (ReLU) layer, and a max-pooling layer with a stride of 2. The layer is connected to repeated residual blocks, as shown by the green blocks in Fig. 1, which consists of convolutional layers, BN layers, and ReLU layers (refer to the original paper [22] for full description).

The decoder comprises repeating upsampling blocks (magenta and purple blocks in Fig. 1) that double the spatial resolutions of the output activations while halving the number of feature channels. Each upsampling block consists of a BN layer, ReLU layer, and a transpose convolution layer of kernel size 2 × 2 and a stride of 2, which provides upsampling. We also add a concurrent spatial and channel squeeze and excitation (SCSE) module [24] between the batch normalization layer and the transpose convolution layer.

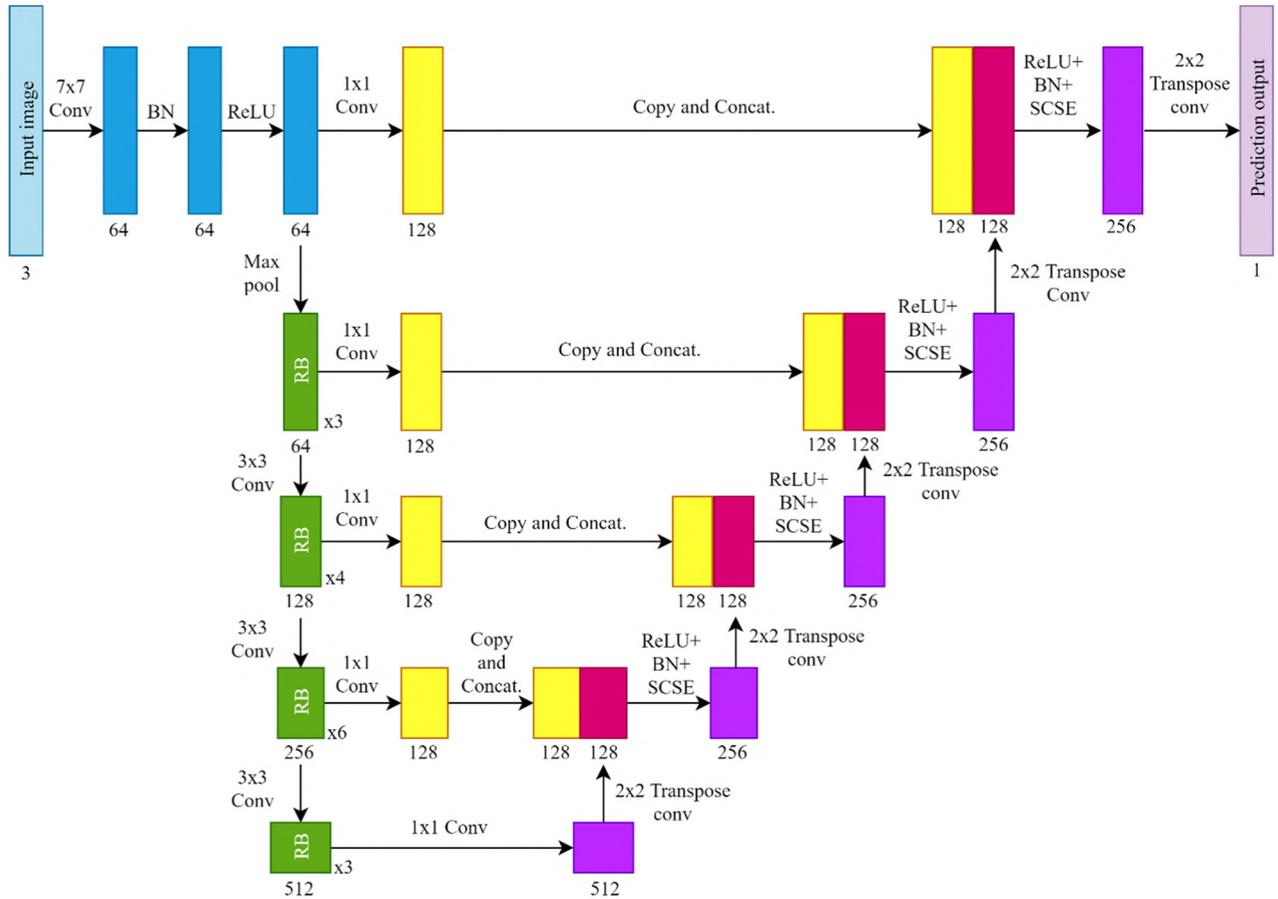

**FIGURE 1.** Proposed network architecture. The blocks represent the activations, whereas the arrows represent the layer operations. The green blocks represent the residual blocks (RB) in ResNet-34. For simplicity of visualization, the magenta blocks represent the activation before passing through successive layers of ReLU, batch normalization (BN), and concurrent spatial and channel squeeze and excitation (SCSE) modules. The number below each block represents the number of feature channels it has.

There are also additional convolution layers that connect several downsampling blocks to their corresponding upsampling blocks with the same spatial resolution to perform a 1 × 1 convolution on each downsampling block. The input of each upsampling block is a channel-wise concatenated tensor of both the output of the previous block and the output activation from the downsampling block. These operations are illustrated by the horizontal arrows connecting the yellow blocks in Fig. 1.

### B. LOSS FUNCTION

We choose the dice coefficient loss [25] as the loss function because it directly optimizes the dice score, which is equivalent to the $F1$ score. We also use this loss function because there is a severe class imbalance. In the pavement crack images, the non-crack pixels outnumber crack pixels by a ratio of around 65:1. We pass the network's output layer through a sigmoid function, so each element in the output layer would have a range of [0,1], which reflects the probability that a crack is present in each pixel.

The following equation determines the dice coefficient loss $L$:

$$L = \frac{1}{N} \sum_{n=1}^{N} 1 - \frac{2|\hat{\boldsymbol{y}} \odot \boldsymbol{y}|}{|\hat{\boldsymbol{y}}| + |\boldsymbol{y}|}, \tag{1}$$

where $\hat{\boldsymbol{y}} \in \mathbb{R}^{h \times w}$ and $\boldsymbol{y} \in \mathbb{R}^{h \times w}$ represent the prediction and ground truth, respectively, the operator $|\cdot|$ denotes the sum of all matrix elements of its argument and $\odot$ denotes the element-wise multiplication operation. The loss is a scalar value that outputs 0 if the prediction completely matches the ground truth, and 1 if otherwise.

### C. PARAMETER OPTIMIZATION

We initialize the parameters in each convolutional layer in the upsampling decoder, which consists of weights and biases, with the method of He et al. [26]. As for the ResNet-34 encoder, we keep the pretrained parameters during initialization.

After initialization, we optimize all neural network parameters with the AdamW [27] optimizer to minimize the loss function. The simplification of the AdamW optimizer is described as follows: First, at iteration $t$, we determine the partial derivative $g_t$ of the loss function $f$ with respect to the parameters of the previous iteration $\theta_{t-1}$:

$$g_t = \nabla_\theta f(\theta_{t-1}). \tag{2}$$

We then compute the first-moment estimate $m_t$ and second-moment estimate $v_t$ as the convex combination of their respective estimates from the previous iteration and the partial derivative of the loss function:

$$m_t = \beta_1 m_{t-1} + (1 - \beta_1) g_t \tag{3}$$

$$v_t = \beta_2 v_{t-1} + (1 - \beta_2) g_t^2, \tag{4}$$

where $\beta_1$ and $\beta_2$ are hyperparameters which we set to 0.9 and 0.999 by default, respectively.

We perform bias correction to reduce the effect of the term $1 - \beta_1$ on $g_t$, and reduce the sensitivity to the initial values $m_0$ and $v_0$, which we arbitrarily set to 0 before the first iteration. The bias correction is done according to

$$\widehat{m}_t = m_t / (1 - \beta_1^t), \tag{5}$$

$$\widehat{v}_t = v_t / (1 - \beta_2^t). \tag{6}$$

Finally, we update the parameters of the next iteration $\theta_t$ using the following equation:

$$\theta_t = (1 - \lambda)\theta_{t-1} - \alpha \left( \frac{\widehat{m}_t}{\sqrt{\widehat{v}_t} + \epsilon} \right) \tag{7}$$

where $\lambda$ is the weight decay (set to 0.01), $\alpha$ is the learning rate, and $\epsilon$ is the epsilon number (set to $10^{-8}$).

### D. TUNING THE LEARNING RATES
We do not fix the learning rate in the optimization procedure above. It changes across the layers in the neural network and with each training iteration, respectively.

#### 1) LEARNING RATES FOR VARIOUS NETWORK LAYERS
We divide the neural network into three groups of layers (*layer-groups,* for short). The first layer-group spans from the input to the 128-channel residual block. The second layer-group spans from the 256-channel residual block to the rest of the encoder. The third layer-group includes the entire upsampling part (the decoder). The three layer-groups receive learning rates of ratio 1/9: 1/3: 1. The rationale for this ratio is that the earlier layers of a convolutional neural network learn features that are more fundamental, such as oriented edges and corners, whereas the latter layers learn more specific features, which are similar to real images [28]. Therefore, when we use the pretrained ResNet-34 for segmentation tasks, we need not train the earlier layers as much as the last layers.

Because we use transfer learning [20] in our network architecture, we fine-tune the neural network in two stages. In the first stage, we "freeze" the first layer-group, i.e., temporarily set the learning rate to zero so that all their parameters are not updated. We train the second and third layer-groups normally for 15 epochs. In the second stage, we "unfreeze" the first layer-group. We continue the training from the first stage for an additional 30 epochs. In other words, we only train the first layer-group after the other two layer-groups are well optimized. By training the latter parts of the pretrained ResNet-34 (second layer-group) and the decoder (third layer-group), this procedure takes maximum advantage of transfer learning. This is because these layers can be optimized for segmentation tasks while keeping the earlier layers of ResNet-34 intact since they are likely to have low-level semantic features.

#### 2) LEARNING RATES ACROSS TRAINING ITERATIONS
Instead of using a constant learning rate throughout all iterations, we use a cyclical learning rate schedule similar to "one-cycle" training, proposed by [29], [30]. We set the minimum learning rate value, $lr_{\min}$, to 5% of the maximum learning rate value, $lr_{\max}$. We then train each batch of images with linearly increasing learning rates, which begin with $lr_{\min}$ and end with $lr_{\max}$ at about 40% of the total number of iterations. Beyond that point, we linearly decrease the learning rates to near zero until the last iteration. We apply the cyclical learning rate schedule only once, regardless of the number of epochs. Note that the values for $lr_{\min}$ and $lr_{\max}$ apply only to the third layer-group. We train the other two layer-groups with different (though similar) cyclical learning rate schedules, but we scale their learning rates by the ratio above of 1/9: 1/3: 1.

## IV. BENCHMARK RESULTS
The specifications of the workstation used to train our neural network are GTX 960M GPU (4 GB VRAM), Intel Core i5 processor, and 16 GB RAM. The deep learning framework used is PyTorch version 0.3.0 [31], and we also used the *fastai* library (version 0.6) [32], which was built on top of PyTorch. (We provide this information so that the reader can calibrate our results appropriately.)

### A. DATASETS
We evaluate our method on the CFD dataset [11] and the Crack500 dataset [17]. We train our neural network on images with progressively increasing sizes: $128 \times 128$, $256 \times 256$, and $320 \times 320$. This is done by resizing and cropping the images. We also performed the image resizing and cropping identically on their corresponding ground truth images to match their corresponding training images.

We apply image augmentations randomly to each image during neural network training to increase the number of training examples virtually. In each training iteration, we perform three types of augmentations:

- Random rotations between $0°$ to $360°$. To deal with corners of rotated images, the pixels near the rectangular corners take the reflection of the rotated image's borders.
- Random flips in horizontal and vertical axes.

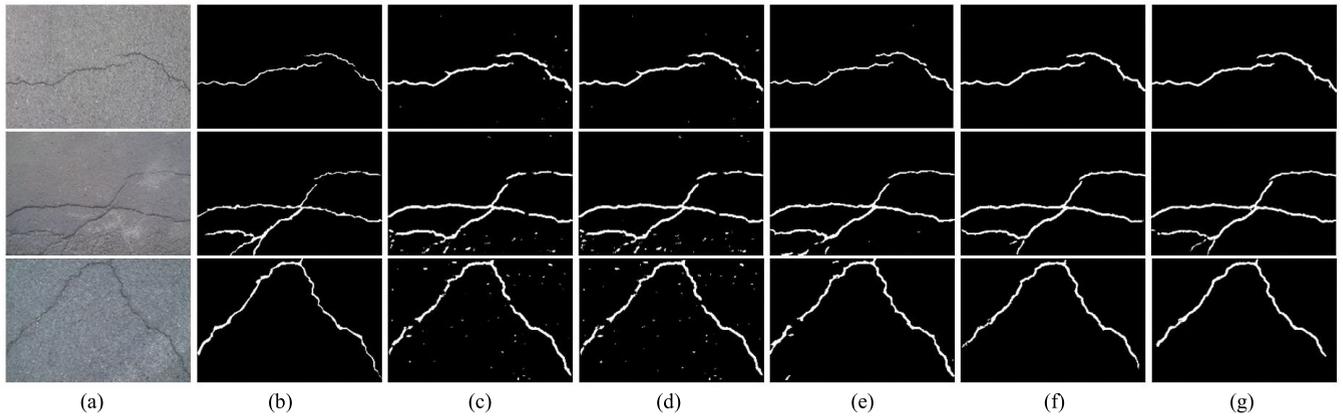

**FIGURE 2.** Comparison of results among various methods on the CFD dataset. From left to right column: (a) original image, (b) ground truth, (c) U-Net by Jenkins *et al.* [18], (d) U-Net by Nguyen *et al.* [19], (e) CNN by Fan *et al.* [14], (f) Split-Attention network [33], (g) our method.

- Random changes in lighting. We randomly increase or decrease the image balance and contrast by a change of 0.05, respectively.

We also apply the same augmentations, except for lighting, to the ground truth of each training image.

### B. RESULTS

To ensure that our neural network can generalize on unseen images, we split the dataset into a training dataset and a test dataset. We only train the neural network on the training dataset and evaluate the performance on the test dataset.

We split the CFD dataset randomly by a 60:40 ratio, resulting in 72 images in the training dataset and 46 images in the test dataset. For the Crack500 dataset, we use the training dataset (3792 images) and test dataset (2248 images) provided by the authors [17].

To get the prediction results, we obtain a binary prediction matrix by setting all elements of $\hat{y}$(refer to (1)) with the value of over 0.5 to be 1 (crack), and 0 (non-crack) otherwise. We select precision ($Pr$), recall ($Re$), and $F1$ score as the performance metrics. Their precise definitions are:

$$Pr = \frac{TP}{TP + FP} \tag{8}$$

$$Re = \frac{TP}{TP + FN} \tag{9}$$

$$F1 = \frac{2 \times Pr \times Re}{Pr + Re} \tag{10}$$

where $TP$, $FP$, and $FN$ are the numbers of true positives, false positives, and false negatives. Note that we do not use the accuracy score as an evaluation metric because the true negatives outweigh the true positives significantly and are therefore easy to predict. Therefore, the number of true negatives does not correctly reflect the quality of the prediction.

Because there are transition regions between the crack pixels and the non-crack pixels in the subjectively labeled ground truth, we consider 2 pixels near any labeled crack pixel as true positives. This evaluation convention was also used in other papers to evaluate their methods on the CFD dataset [11]. We also adopt this evaluation convention on the Crack500 dataset [17]. We have taken the average $Pr$, $Re$, and $F1$ scores for all test images.

**TABLE 1.** Comparison of various methods on CFD dataset.

| Method | $Pr$ | $Re$ | $F1$ |
|---|---|---|---|
| U-Net by Jenkins et al. [18] | 0.8517 | 0.9155 | 0.8727 |
| U-Net by Nguyen et al. [19] | 0.8567 | 0.9132 | 0.8745 |
| CNN by Fan et al. [14] | 0.9119 | **0.9481** | 0.9244 |
| Split-Attention Network [33] | 0.9701 | 0.9375 | 0.9521 |
| Our method | **0.9702** | 0.9432 | **0.9555** |

**TABLE 2.** Comparison of various methods on Crack500 dataset.

| Method | $Pr$ | $Re$ | $F1$ |
|---|---|---|---|
| U-Net by Jenkins et al. [18] | 0.6811 | 0.6629 | 0.6788 |
| U-Net by Nguyen et al. [19] | 0.6954 | 0.6744 | 0.6895 |
| CNN by Fan et al. [14] | 0.7123 | 0.6955 | 0.7056 |
| Split-Attention Network [33] | 0.7368 | 0.7165 | 0.7295 |
| Our method | **0.7426** | **0.7285** | **0.7327** |

Table 1 shows a quantitative comparison of the precision, recall, and $F1$ scores for our proposed method and four other recent segmentation algorithms we have reimplemented. (We have highlighted the best $Pr$, $Re$, and $F1$ scores in bold.) From Table 1, we can see that our proposed method achieves higher precision than the other compared methods. Although the recall is marginally lower than the method of Fan *et al.* [14], our proposed method achieves an overall $F1$ score of 95.55%, higher than the other methods (see Table 1). Besides, our method also gets the highest $F1$ score (73.27%) on the Crack500 dataset (see Table 2). In both datasets, our method marginally outperforms the Split-Attention Network [33], which uses a significantly larger neural network. For visual comparison, we have included a few prediction results by the on the test dataset of CFD and Crack500 in Fig. 2 and Fig. 3, respectively.

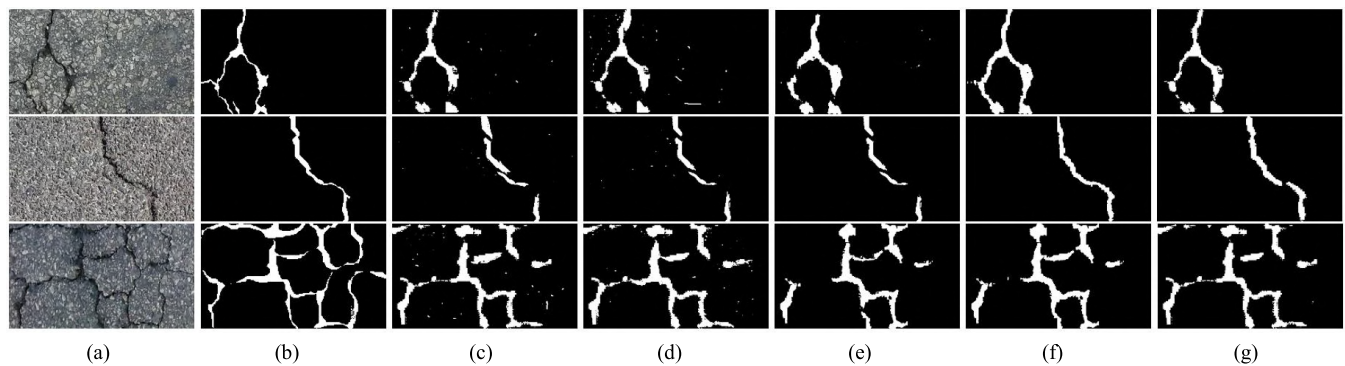

**FIGURE 3.** Comparison of results among various methods on the Crack500 dataset. From left to right column: (a) original image, (b) ground truth, (c) U-Net by Jenkins *et al.* [18], (d) U-Net by Nguyen *et al.* [19], (e) CNN by Fan *et al.* [14], (f) Split-Attention Network [33], (g) our method.

## V. ABLATION STUDIES

We perform ablation studies on the CFD dataset to show the improvements in the performance due to our algorithm design choices.

### A. COMPARISON BETWEEN ONE-STAGE TRAINING AND TWO-STAGE TRAINING

The term *one-stage training* refers to the standard training procedure, where we train the neural network without freezing some layers. The term *two-stage training* refers to the training procedure where we first train the neural network with some layers frozen, then train it with all layers unfrozen. We compare the performances of one-stage training and two-stage training. In one-stage training, we train the neural network for 30 epochs, identical to the second stage of the two-stage training. To make this a fair comparison, we do not add the concurrent spatial and channel squeeze and excitation (SCSE) blocks to the decoder in both cases, and we use only $320 \times 320$ images for training.

The experimental results (see Table 3) show that two-stage training gives better performance than one-stage training. In [20], Yosinski *et al.* tested the idea of freezing transferred layers and fine-tuning them. However, they only concluded that fine-tuning is better than freezing the layers. In this paper, we show that we can get better performance by combining the two procedures.

### B. THE ADDITION OF CONCURRENT SPATIAL AND CHANNEL SQUEEZE AND EXCITATION (SCSE) MODULES

We train two neural networks in this experiment, one with concurrent spatial and channel squeeze and excitation (SCSE) modules, and the other without the SCSE modules. We train both neural networks using two-stage training, and only on $320 \times 320$ images. We keep the other procedures the same. From Table 4, we can see that the SCSE modules improve the result by a small margin, at the expense of the recall score, while increasing the precision and $F1$ score.

The authors who proposed the SCSE modules [24] suggested inserting these modules in both the encoder and decoder of a fully convolutional network such as U-Net.

**TABLE 3.** Performance comparison of one-stage training and two-stage training.

|  | *Pr* | *Re* | *F1* |
|---|---|---|---|
| One-stage | 0.9671 | 0.9344 | 0.9492 |
| Two-stage | 0.9683 | 0.9423 | 0.9542 |

**TABLE 4.** Performance comparison between architectures with and without SCSE modules.

|  | *Pr* | *Re* | *F1* |
|---|---|---|---|
| Without SCSE | 0.9683 | 0.9423 | 0.9542 |
| With SCSE | 0.9724 | 0.9390 | 0.9546 |

However, since our encoder is a pretrained neural network, there are co-adapted interactions between the layers in convolutional neural nets [20]. Hence, we do not add new layers that are randomly initialized in between the pre-existing layers. Nonetheless, by only adding the SCSE modules in the decoder part of the network, we observe an increase in the performance.

### C. TRAINING WITH INCREASING IMAGE SIZES

Since each crack pattern is random in pavement images, we can extract smaller images from each of the $320 \times 480$ images. We compare the performances of the neural network after it has been:

a) Trained with image sizes of $128 \times 128$, $256 \times 256$, and then $320 \times 320$
b) Trained with image sizes $320 \times 320$ only.

To make a fair comparison, we train the neural network for 30 epochs for each image size in case (a), and 90 epochs for case (b). Table 5 shows the comparison between the training procedures.

From Table 5, all metrics have improved after training with progressively increasing image sizes. This is because the parameters in the neural network that have been trained on the smaller sizes provide proper initialization for the training on the images of size $320 \times 320$. On the other hand, the neural network that is only trained on the $320 \times 320$ images alone did not have the advantage. We also hypothesize that the variation of images the neural network has seen contributes to the increase in performance. Interestingly, even though the smaller training images of size $128 \times 128$ and $256 \times 256$ fed to the neural network are subsets of the $320 \times 320$ images, the neural network can generalize better.

**TABLE 5.** Performance comparison when training with only one image size and three different image sizes.

| Image Size | Pr | Re | F1 |
|---|---|---|---|
| Only 320×320 | 0.9683 | 0.9423 | 0.9542 |
| All image sizes | 0.9698 | 0.9431 | 0.9549 |

## VI. CONCLUSION

Computer algorithms are often used to automate segmentation tasks for pavement crack images. We have shown that our deep learning technique can solve pavement crack segmentation tasks accurately. Our network architecture is a U-Net with an encoder of pretrained ResNet-34. We deploy the "one-cycle" training schedule to speed up the convergence. We also adopt techniques such as freezing layer-groups, assigning different learning rates to each layer-group, and increasing image sizes progressively. Our approach achieves an $F1$ score of about 96% on the CFD dataset, and about 73% on the Crack500 dataset. Our approach requires minimal feature engineering compared to other machine learning techniques; hence, it is useful when there is a lack of domain expertise in analyzing pavement cracks.

One limitation of this study is that our algorithm requires every pixel of ground truth images to be manually labeled, making data acquisition expensive. One research direction to mitigate this issue is unsupervised learning-based techniques that do not require any ground truth labels. Since supervised learning algorithms aim to fit the function approximator to the given labeled training data, the performance of such algorithms on unseen real-life data largely depends on how much the training dataset reflects the real-life pavement crack images. Hence, practical algorithm deployment potentially requires a much bigger training dataset than the ones we use to capture a wider data distribution. Further improvements can be made by collecting more high-resolution training images and high-quality ground truth annotations. Besides, more studies can be done to discover more augmentation techniques to alleviate the problem of small datasets.

## REFERENCES

[1] S. S. Adlinge and A. K. Gupta, "Pavement deterioration and its causes," in *Proc. IOSR JMCE*, Apr. 2013, pp. 9–15.

[2] E. Zaloshnja and T. R. Miller, "Cost of crashes related to road conditions, United States, 2006," *Ann. Adv. Automot. Med.*, vol. 53, pp. 53–141, Oct. 2009.

[3] K. C. Wang, Z. Hou, and W. Gong, "Automation techniques for digital highway data vehicle (DHDV)," presented at the 7th Int. Conf. Manag. Pavement Assets, 2008.

[4] L. Sjogren and P. Offrell, "Automatic crack measurement in Sweden," in *Proc. 4th Int. Symp. Pavement Surface Characteristics Roads Airfields World Road Assoc. (PIARC)*, 2000, pp. 497–506.

[5] G. Caroff, P. Joubert, F. Prudhomme, and G. Soussain, "Classification of pavement distresses by image processing (MACADAM SYSTEM)," in *Proc. ASCE*, 1989, pp. 46–51.

[6] H. Oliveira and P. Correia, "Automatic road crack segmentation using entropy and image dynamic thresholding," in *Proc. EUSIPCO*, Glasgow, U.K., Aug. 2009, pp. 622–626.

[7] N. Tanaka and K. Uematsu, "A crack detection method in road surface images using morphology," in *Proc. MVA*, Chiba, Japan, 1998, pp. 154–157.

[8] H. Zhao, G. Qin, and X. Wang, "Improvement of canny algorithm based on pavement edge detection," in *Proc. 3rd Int. Congr. Image Signal Process.*, Oct. 2010, pp. 964–967.

[9] K. Wang, Y. Wang, J. O. Strandhagen, and T. Yu, *Advanced Manufacturing and Automation VII*. Singapore: Springer, 2018.

[10] D. Ai, G. Jiang, L. Siew Kei, and C. Li, "Automatic pixel-level pavement crack detection using information of multi-scale neighborhoods," *IEEE Access*, vol. 6, pp. 24452–24463, 2018.

[11] Y. Shi, L. Cui, Z. Qi, F. Meng, and Z. Chen, "Automatic road crack detection using random structured forests," *IEEE Trans. Intell. Transp. Syst.*, vol. 17, no. 12, pp. 3434–3445, Dec. 2016.

[12] H. Li, D. Song, Y. Liu, and B. Li, "Automatic crack detection and classification by multi-scale image fusion," *IEEE Trans. Intell. Transp. Syst.*, vol. 20, no. 6, pp. 2025–2036, Jun. 2019.

[13] S. Bang, S. Park, H. Kim, Y.-S. Yoon, and H. Kim, "A deep residual network with transfer learning for pixel-level road crack detection," in *Proc. ISARC*, vol. 35, 2018, pp. 1–4.

[14] Z. Fan, Y. Wu, J. Lu, and W. Li, "Automatic pavement crack detection based on structured prediction with the convolutional neural network," 2018, *arXiv:1802.02208*. [Online]. Available: http://arxiv.org/abs/1802.02208

[15] A. Zhang, K. C. P. Wang, B. Li, E. Yang, X. Dai, Y. Peng, Y. Fei, Y. Liu, J. Q. Li, and C. Chen, "Automated pixel-level pavement crack detection on 3D asphalt surfaces using a deep-learning network," *Comput.-Aided Civil Infrastruct. Eng.*, vol. 32, no. 10, pp. 805–819, Oct. 2017.

[16] A. Zhang, K. C. P. Wang, Y. Fei, Y. Liu, C. Chen, G. Yang, J. Q. Li, E. Yang, and S. Qiu, "Automated pixel-level pavement crack detection on 3D asphalt surfaces with a recurrent neural network," *Comput.-Aided Civil Infrastruct. Eng.*, vol. 34, no. 3, pp. 213–229, Mar. 2019.

[17] F. Yang, L. Zhang, S. Yu, D. Prokhorov, X. Mei, and H. Ling, "Feature pyramid and hierarchical boosting network for pavement crack detection," *IEEE Trans. Intell. Transp. Syst.*, vol. 21, no. 4, pp. 1525–1535, Apr. 2020.

[18] M. David Jenkins, T. A. Carr, M. I. Iglesias, T. Buggy, and G. Morison, "A deep convolutional neural network for semantic pixel-wise segmentation of road and pavement surface cracks," in *Proc. 26th Eur. Signal Process. Conf. (EUSIPCO)*, Sep. 2018, pp. 2120–2124.

[19] A. Sheta, H. Turabieh, S. Aljahdali, and A. Alangari, "Pavement crack detection using convolutional neural network," in *Proc. 9th Int. Symp. Inf. Commun. Technol.*, 2018, pp. 251–256.

[20] J. Yosinski, J. Clune, Y. Bengio, and H. Lipson, "How transferable are features in deep neural networks," in *Proc. NIPS*, Montreal, QC, Canada, Dec. 2016, pp. 3320–3328.

[21] O. Ronneberger, P. Fischer, and T. Brox, "U-Net: Convolutional networks for biomedical image segmentation," in *Proc. MICCAI*, Munich, Germany, Oct. 2017, pp. 234–241.

[22] K. He, X. Zhang, S. Ren, and J. Sun, "Deep residual learning for image recognition," in *Proc. IEEE Conf. Comput. Vis. Pattern Recognit. (CVPR)*, Jun. 2016, pp. 770–778.

[23] S. Ioffe and C. Szegedy, "Batch normalization: Accelerating deep network training by reducing internal covariate shift," in *Proc. ICML*, Lille, France, Jul. 2015, pp. 448–456.

[24] A. G. Roy, N. Navab, and C. Wachinger, "Concurrent spatial and channel squeeze & excitation in fully convolutional networks," in *Proc. MICCAI*, Granada, Spain, 2018, pp. 421–429.

[25] F. Milletari, N. Navab, and S.-A. Ahmadi, "V-Net: Fully convolutional neural networks for volumetric medical image segmentation," 2016, *arXiv:1606.04797*. [Online]. Available: http://arxiv.org/abs/1606.04797

[26] K. He, X. Zhang, S. Ren, and J. Sun, "Delving deep into rectifiers: Surpassing human-level performance on ImageNet classification," in *Proc. IEEE Int. Conf. Comput. Vis. (ICCV)*, Dec. 2015, pp. 1026–1034.

[27] I. Loshchilov and F. Hutter, "Decoupled weight decay regularization," presented at the ICLR, New Orleans, LA, USA, 2019. [Online]. Available: https://openreview.net/forum?id=Bkg6RiCqY7

[28] M. D. Zeiler and R. Fergus, "Visualizing and understanding convolutional networks," in *Proc. Eur. Conf. Comput. Vis.*, 2013, pp. 818–833.

[29] L. N. Smith, "Cyclical learning rates for training neural networks," in *Proc. IEEE Winter Conf. Appl. Comput. Vis. (WACV)*, Mar. 2017, pp. 464–472.

[30] L. N. Smith, "A disciplined approach to neural network hyper-parameters: Part 1—Learning rate, batch size, momentum, and weight decay," 2018, *arXiv:1803.09820*. [Online]. Available: https://arxiv.org/abs/1803.09820

[31] A. Paszke, S. Gross, F. Massa, A. Lerer, J. Bradbury, G. Chanan, T. Killeen, Z. Lin, N. Gimelshein, L. Antiga, and A. Desmaison, "PyTorch: An imperative style, high-performance deep learning library," in *Proc. NeurIPS*, Vancouver, BC, Canada, 2019, pp. 8024–8035.

[32] J. Howard and S. Gugger, "Fastai: A layered API for deep learning," 2020, *arXiv:2002.04688*. [Online]. Available: http://arxiv.org/abs/2002.04688

[33] H. Zhang, C. Wu, Z. Zhang, Y. Zhu, Z. Zhang, H. Lin, Y. Sun, T. He, J. Mueller, R. Manmatha, M. Li, and A. Smola, "ResNeSt: Split-attention networks," 2020, *arXiv:2004.08955*. [Online]. Available: http://arxiv.org/abs/2004.08955

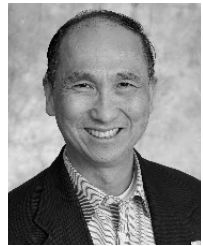

**EDWIN K. P. CHONG** (Fellow, IEEE) received the B.E. degree (Hons.) from The University of Adelaide, Australia, and the M.A. and Ph.D. degrees from Princeton University, where he held an IBM Fellowship, in 1991. He joined the School of Electrical and Computer Engineering (ECE), Purdue University, in 1991, where he was named as a University Faculty Scholar, in 1999. Since August 2001, he has been a Professor of ECE and mathematics with Colorado State University. He has coauthored the best-selling book, *An Introduction to Optimization* (Fourth Edition, Wiley-Interscience, 2013). He was the President of the IEEE Control Systems Society, in 2017.

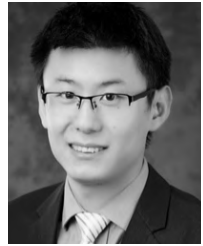

**XU YANG** (Member, IEEE) received the bachelor's degree from Southeast University, China, in 2009, and the Ph.D. degree from Michigan Technological University, USA, in 2015. He is currently a Lecturer with the Department of Civil Engineering, Monash University, Australia. His research interests include advanced road pavement construction and maintenance technology, deep learning for automated pavement distress detection, and numerical simulation for civil engineering materials.

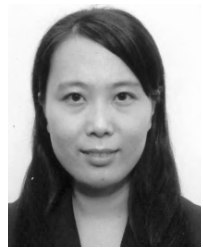

**XIN WANG** (Senior Member, IEEE) received the Ph.D. degree from Nanyang Technological University (NTU), in 2007. She was a Research Fellow with the Robotics Research Centre, NTU, from 2007 to 2009. She was an Assistant Professor with the Faculty of Engineering and Science, Universiti Tunku Abdul Rahman, from 2009 to 2012. She is currently an Associate Professor with the School of Engineering, Monash University Malaysia. Her research interests include optical metrology, 3-D imaging, nondestructive inspection, and machine vision.

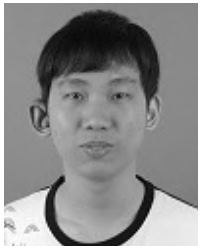

**STEPHEN L. H. LAU** was born in Sibu, Malaysia, in 1995. He received the B.S. degree in mechanical engineering from Monash University Malaysia, in 2019. His current research interests include computer vision, generative models, and reinforcement learning.

• • •